%% file: main.tex
\documentclass[runningheads]{llncs}

\usepackage{accv}

\usepackage{accvabbrv}

\usepackage{graphicx}
\usepackage{booktabs}
\usepackage{fontawesome5}

\usepackage[accsupp]{axessibility}  % Improves PDF readability for those with disabilities.

\usepackage{hyperref}

\usepackage{orcidlink}

\begin{document}

\title{MuViSeg: Multi-View Segment Correspondences from Dense Geometry
Priors} 

\titlerunning{MuViSeg}

\author{Denis Fatykhoph\orcidlink{0009-0004-3998-1915} \and
Timur Akhtyamov\orcidlink{0000-0003-0877-3318} \and
Konstantin Pakulev\orcidlink{0000-0001-7627-0400} \and
German Devchich\orcidlink{0009-0003-8000-4342} \and
Gonzalo Ferrer\orcidlink{0000-0003-2704-7186}}

\authorrunning{D.~Fatykhoph et al.}

\institute{Applied AI Institute, Moscow, Russia\\
\email{d.fatykhoph@applied-ai.ru} \\[1.6ex]
\faGithub\;Project Page: \url{https://muviseg.github.io/}}

\maketitle

\begin{abstract}
    Object-level mapping and topological navigation reason about whole objects, yet the correspondences they rely on are computed over keypoints or pixels and only aggregated into objects afterwards. A recent line of work removes this detour by matching directly at the level of instance segments: a class-agnostic segmenter partitions each image, and per-segment descriptors are pooled from large 3D foundation models over the masks. This shifts the open question to how frozen foundation features should be processed once the unit of matching is a segment. We introduce MuViSeg, three learned matching heads: a LightGlue-style attention head on frozen MASt3R descriptors; a DPT-style multi-scale fusion module that exposes layered spatial detail from VGGT before pooling; and, as our main contribution, a joint multi-view head that attends over segments from several views at once, recovering transitive correspondences that pairwise matchers cannot express. In zero-shot evaluation on Replica and Virtual KITTI 2 across 0°–180° viewpoint changes, our heads consistently improve over a parameter-free Sinkhorn matcher on the same backbone. Across 102 HM3D navigation episodes, direct segment matchers and sparse keypoint aggregation are statistically indistinguishable, while dense matches voted into masks lose 13.7–18.0 SPL.
\end{abstract}

\input{sections/intro}
\input{sections/related_work}
\input{sections/materials_and_methods}
\input{sections/experiments}
\input{sections/conclusions}
\input{sections/acknowledgements}

% \clearpage\mbox{}Page \thepage\ of the manuscript.
% \clearpage\mbox{}Page \thepage\ of the manuscript.
% \clearpage\mbox{}Page \thepage\ of the manuscript.
% \clearpage\mbox{}Page \thepage\ of the manuscript.
% \clearpage\mbox{}Page \thepage\ of the manuscript. This is the last page.
% Now we have reached the maximum length of an ACCV \ACCVyear{} submission (excluding references).
% References should start immediately after the main text, but can continue past p.\ 14 if needed.
% \clearpage  % TODO REVIEW/FINAL: This \clearpage needs to be removed from both review and camera-ready versions.

\bibliographystyle{splncs04}
\bibliography{main}

\end{document}

%% file: sections/intro.tex
\section{Introduction} \label{sec:introduction}

Image correspondence underpins object tracking~\cite{ravi2024sam2}, topological navigation~\cite{sarkar2024robohop}, and scene-graph construction~\cite{li2022sgtr, hughes2022hydra}. These systems ultimately require \emph{object identities}, yet most matchers return sparse keypoint or dense pixel correspondences~\cite{detone2018superpoint, sarlin2020superglue, lindenberger2023lightglue, sun2021loftr, edstedt2024roma}. Recovering object associations then requires voting low-level matches inside segmentation masks, introducing an additional failure point. The problem is especially difficult under wide baselines, where viewpoint, scale, and visible object surfaces change substantially.

A direct alternative is to match instance segments themselves: SAM~\cite{kirillov2023sam, ravi2024sam2} partitions each image, and backbone features are pooled into one descriptor per mask. RoboHop~\cite{sarkar2024robohop} uses appearance features~\cite{oquab2023dinov2}, while recent methods draw on 3D foundation models~\cite{wang2024dust3r, leroy2024mast3r, wang2025vggt}. SegMASt3R~\cite{sarkar2025segmastr}, our closest baseline, learns $24$-dim segment descriptors from frozen MASt3R patch features and matches them with a parameter-free Sinkhorn layer~\cite{sinkhorn1967concerning, sarlin2020superglue}.

We ask how frozen foundation features should be processed once the matching unit is a segment. We test three hypotheses: \textbf{(H1)} place learned capacity in a LightGlue-style~\cite{lindenberger2023lightglue} cross-segment head on frozen MASt3R descriptors; \textbf{(H2)} preserve spatial detail through DPT-style~\cite{ranftl2021dpt} multi-scale VGGT fusion before mask pooling; and \textbf{(H3)} score segments from multiple views \emph{jointly}. Our main method addresses H3 by concatenating segments from $N$ views and applying shared self-attention with per-view embeddings, allowing intermediate views to support correspondences unavailable to isolated pairwise matching.

We train on ScanNet++~\cite{yeshwanth2023scannetpp} and evaluate zero-shot on Replica~\cite{straub2019replica} and Virtual KITTI~2~\cite{cabon2020vkitti2} across $0^\circ$--$180^\circ$ viewpoint changes. The pairwise MASt3R head is strongest overall and outdoors, whereas joint attention helps the VGGT head most at wide baselines; the Sinkhorn baseline remains competitive at extreme rotations. In $102$ HM3D navigation episodes~\cite{krantz2023iin, ramakrishnan2021hm3d}, direct segment matchers and sparse keypoint aggregation are not statistically separated, while RoMa's dense pixel votes underperform every direct matcher.

\paragraph{Contributions.}
(i) A joint multi-view segment matcher that generalises to tuple sizes unseen during training.
(ii) A stratified zero-shot comparison identifying which segment-processing pipeline works best in each viewpoint regime.
(iii) A $102$-episode closed-loop evaluation with paired significance testing: direct matchers outperform the tested dense-vote baseline but are not separated from sparse aggregation.

%% file: sections/related_work.tex
\section{Related Work} \label{sec:related_work}

\paragraph{Sparse and dense feature matching.}
Sparse pipelines~\cite{detone2018superpoint, sarlin2020superglue, lindenberger2023lightglue} produce keypoint-level matches; dense methods~\cite{sun2021loftr, edstedt2024roma, edstedt2023dkm} predict per-pixel correspondences via 4D correlation volumes or kernelised regression, often on top of frozen self-supervised features~\cite{oquab2023dinov2}, with GIM~\cite{xuelun2024gim} addressing distribution shift through internet-video self-training. LightGlue~\cite{lindenberger2023lightglue} streamlines the matcher through alternating self- and cross-attention, a DoubleSoftmax scorer, and a learned matchability classifier --- a design we lift to the segment level. None of these methods produce object-level associations directly: a downstream system must aggregate sub-object matches inside a mask, and the resulting associations are only as reliable as the voting rule allows.

\paragraph{3D foundation models.}
DUSt3R~\cite{wang2024dust3r} casts stereo reconstruction as regression of per-pixel pointmaps; MASt3R~\cite{leroy2024mast3r} extends it with prediction heads --- the cross-view decoder outputs patch features of shape $(H/16, W/16, 768)$, and a local-feature head additionally yields $24$-dim per-patch descriptors for pixel-level matching. MASt3R encodes the two images independently and exchanges information through a CroCo-style~\cite{weinzaepfel2023croco} cross-view decoder; its features are inherently \emph{pair-dependent}. VGGT~\cite{wang2025vggt} takes a contrasting route: a $1$B-parameter ViT encoder followed by a $24$-layer Aggregator that processes patch tokens from \emph{all} input views in a single shared sequence, with cross-view interaction emerging from self-attention rather than explicit cross-attention. This contrast --- pairwise cross-attention vs.\ joint multi-view self-attention --- dictates which heads are architecturally compatible with which backbone, and motivates the two head designs we contrast.

\paragraph{Segment-level matching.}
RoboHop~\cite{sarkar2024robohop} builds a topological map from SAM masks, pools DINOv2~\cite{oquab2023dinov2} features over each mask, and matches by cosine similarity --- without explicit multi-view geometry or any learned matching component; TANGO~\cite{podgorski2025tango} extends this segment-based topological navigation with local metric control, but likewise treats segment association as a fixed feature-similarity step rather than a learned matcher. SegMASt3R~\cite{sarkar2025segmastr} is the most directly comparable prior work: it freezes MASt3R's decoder, takes its $768$-dim patch features, and learns a \emph{segment-feature head} (an upsampling MLP) producing $24$-dim per-segment descriptors, matched by a parameter-free Sinkhorn~\cite{sinkhorn1967concerning, sarlin2020superglue} solver with a learnable dustbin logit. All learnable capacity sits in the per-segment feature head; the matcher itself does no cross-segment mixing. We instead place capacity in a cross-segment head --- a wider projected space with explicit self-/cross-attention --- and additionally exploit VGGT's pair-independent Aggregator to open joint attention across more than two views, a design axis inaccessible to MASt3R-based matchers.

\paragraph{Area-based matching.}
A parallel line uses area-level correspondence as an intermediate step for point matching: MESA~\cite{zhang2024mesa, zhang2026mesa} and A2PM~\cite{zhang2026a2pm} establish area matches to narrow the search for keypoints and are ultimately scored by the pose those keypoints yield, so the areas are discarded downstream. MASA~\cite{li2024masa} associates instances across video frames, where temporal continuity constrains the association. Our setting differs in that segments \emph{are} the output: the topological map stores identities, we score them by instance-ID AUPRC/R@1, and the two views of an object may be separated by a wide baseline rather than a frame interval. The two directions are complementary.

%% file: sections/materials_and_methods.tex
\section{Method} \label{sec:method}

% Keep wide figures at the top of a text page rather than on a float-only page.
\renewcommand{\dbltopfraction}{0.95}
\renewcommand{\textfraction}{0.05}
\renewcommand{\dblfloatpagefraction}{0.95}

We address segment matching with a unified template: a frozen vision backbone $\Phi$ produces per-patch features, a class-agnostic segmenter partitions each image into instance masks, masked average pooling produces one descriptor per segment, and a head refines the descriptors and predicts correspondences. Within this template we vary three design axes that follow the openings identified in Section~\ref{sec:introduction}: the \emph{placement of learnable capacity} (in the per-segment feature head before a parameter-free solver, as in SegMASt3R~\cite{sarkar2025segmastr}; or in a learned cross-segment attention head on top of frozen per-segment descriptors, as in our work), the \emph{choice of backbone} (MASt3R vs.\ VGGT), and the \emph{breadth of context} (a single pair vs.\ a tuple of jointly attended views).

\begin{figure*}[!t]
  \centering
  \includegraphics[width=0.85\linewidth]{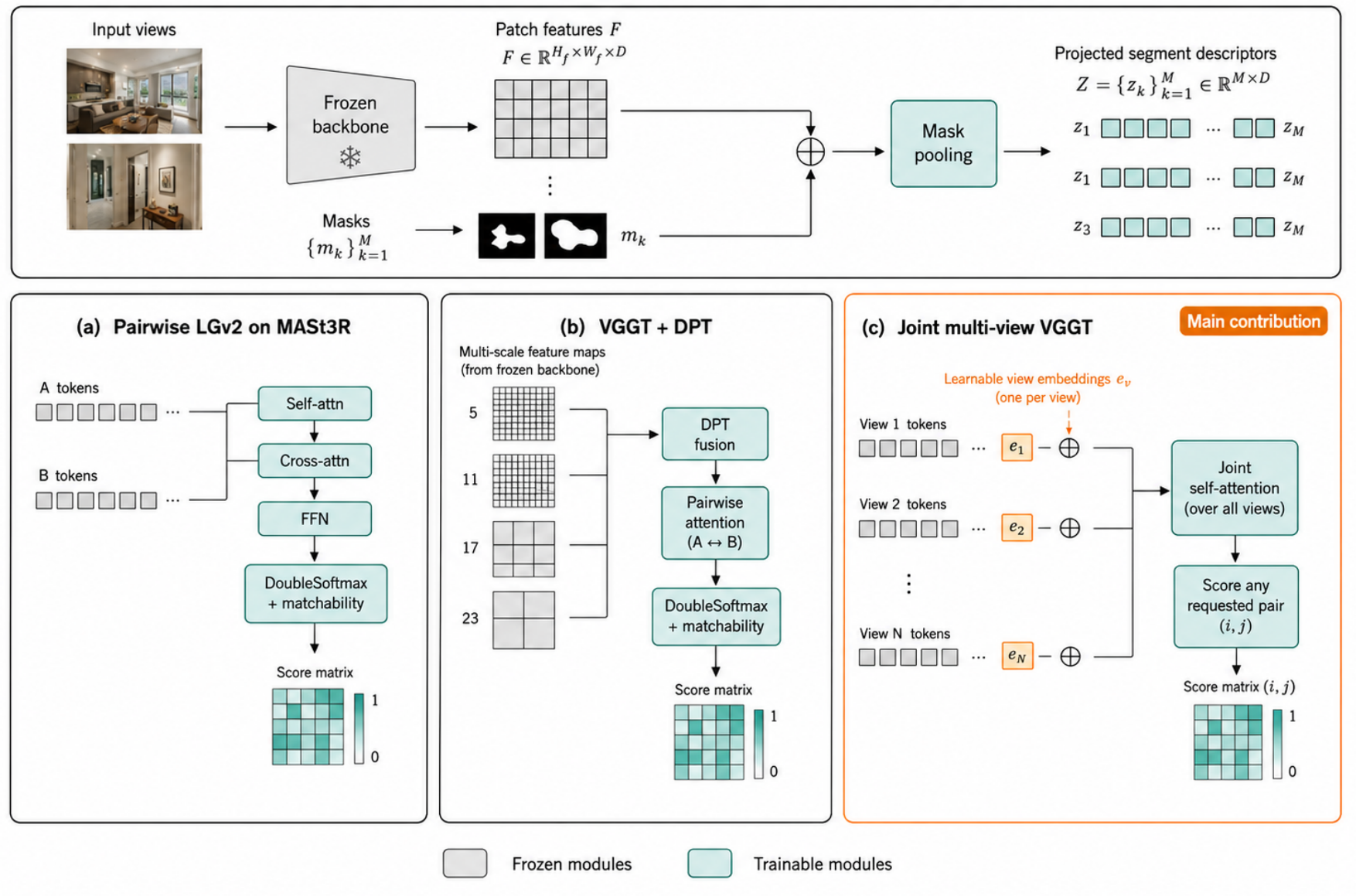}
  \caption{Architecture of the three matchers (grey: frozen, teal: trainable). \textbf{Top:} shared template --- frozen backbone $\to$ per-patch features $\mathbf{F}$, mask-selected and mask-pooled to one descriptor $\mathbf{z}_k$ per segment. The variants differ in feature tap, spatial fusion, and attention context. \textbf{(a)~LGv2 on MASt3R:} pairwise self-/cross-attention and an FFN, then a DoubleSoftmax scorer with a matchability head. \textbf{(b)~VGGT+DPT:} a DPT fusion of Aggregator layers $\{5,11,17,23\}$ before pooling; head as in~(a). \textbf{(c)~Joint multi-view VGGT} (main contribution): tokens from all $N$ views, tagged with per-view embeddings $\mathbf{e}_v$, share one joint self-attention, after which the scorer reads off any pair $(i,j)$.}
  \label{fig:architecture}
\end{figure*}

\paragraph{Why two backbones.}
Our object of study is the matching head, yet we deliberately vary the backbone too, because the backbone is not an interchangeable feature extractor here: it determines \emph{which heads are architecturally possible}. MASt3R's \emph{pair-dependent} features suit a strong pairwise head but cannot be shared across a variable set of views, whereas VGGT's \emph{pair-independent} features are exactly what a joint multi-view head requires (\S\ref{sec:backbones}). The central axis of this work, single-pair versus joint multi-view matching, is therefore only reachable on VGGT; the backbone comparison is a consequence of studying that axis, and we compare head designs under a fixed backbone wherever possible (DPT vs.\ Joint on VGGT); LGv2 vs.\ Sinkhorn also holds MASt3R fixed but changes the feature tap as well as the head (Section~\ref{sec:heads}). The three models --- \emph{SegMASt3R+LGv2}, \emph{SegVGGT-DPT}, and \emph{SegVGGT-DPT Joint} (Fig.~\ref{fig:architecture}) --- share the scaffolding below.

% Declare the wide overview only after enough Method text has been emitted to
% close the page containing the end of Related Work. Otherwise figure* may
% legally jump to the top of that preceding page.

\subsection{Problem formulation} \label{sec:problem}

Let $I_A, I_B \in \mathbb{R}^{H \times W \times 3}$ be two views of the same scene. A class-agnostic segmentation model (SAM~\cite{kirillov2023sam} in our experiments) produces non-overlapping instance masks $\mathcal{M}_A = \{m_1^A, \ldots, m_M^A\}$, $\mathcal{M}_B = \{m_1^B, \ldots, m_N^B\}$. Mask counts $M$ and $N$ are capped at $M_{\max}$, the maximum number of segments per image. A frozen backbone $\Phi$ produces per-patch features $\mathbf{F}_A, \mathbf{F}_B = \Phi(I_A, I_B) \in \mathbb{R}^{H_f \times W_f \times D_{\text{raw}}}$ with raw feature dimension $D_{\text{raw}}$, grid dimensions $(H_f, W_f) = (H/p, W/p)$, and patch size $p$ ($p = 16$ for MASt3R, $p = 14$ for VGGT); $\mathbf{F}_A$ may or may not depend on $I_B$ depending on the backbone (Section~\ref{sec:backbones}). Per-segment descriptors are obtained by masked average pooling (masks resized via nearest-neighbour),
\begin{equation}
  \mathbf{d}_k^v = \frac{1}{|m_k^v|} \sum_{i \in m_k^v} \mathbf{f}_i^v,
  \label{eq:pool}
\end{equation}
where $v\in\{A,B\}$ indexes the view, $\mathbf{f}_i^v$ is the feature at patch $i$, and $|m_k^v|$ is the number of patches selected by mask $m_k^v$. A learnable projector $\pi_\theta : \mathbb{R}^{D_{\text{raw}}} \to \mathbb{R}^D$, with parameters $\theta$ and output dimension $D$, maps them to a shared embedding $\mathbf{z}_k^v = \pi_\theta(\mathbf{d}_k^v)$.

\paragraph{Matching head.}
The head $g_\theta : (\{\mathbf{z}_k^A\}, \{\mathbf{z}_k^B\}) \mapsto (\mathbf{S}, \{\sigma_k^A\}, \{\sigma_k^B\})$ returns a score matrix $\mathbf{S} \in \mathbb{R}^{M \times N}$ and per-segment matchability logits. Internally, $g_\theta$ refines descriptors through self- and cross-attention --- letting each segment attend to others in the same image and in the other view --- yielding $\tilde{\mathbf{z}}_k^v \in \mathbb{R}^D$. Let $\tau>0$ denote the softmax temperature. From the scaled affinity $A_{ij} = \tilde{\mathbf{z}}_i^A \cdot \tilde{\mathbf{z}}_j^B / (\tau \sqrt{D})$ we form two conditional distributions: a row-softmax normalising each row of $\mathbf{A}$ over the columns, and a column-softmax normalising each column over the rows,
\begin{equation}
  p(j \mid i) = \frac{\exp A_{ij}}{\sum_{j'} \exp A_{ij'}},
  \qquad
  p(i \mid j) = \frac{\exp A_{ij}}{\sum_{i'} \exp A_{i'j}}.
  \label{eq:softmax-cond}
\end{equation}
The \emph{DoubleSoftmax} score is the log of their product,
\begin{equation}
  S_{ij} = \log p(j \mid i) + \log p(i \mid j),
  \label{eq:doublesoftmax}
\end{equation}
which is large only when segment $j$ is the best match for $i$ \emph{and} $i$ is the best match for $j$ --- \ie{} it rewards mutually most-similar pairs.

\paragraph{Matchability and the dustbin.}
Not every segment has a counterpart in the other view --- some are occluded, out of frame, or simply absent --- so the matcher must be able to abstain. Each segment carries a \emph{matchability logit} $\sigma_k^v$, predicted by a small classifier on top of $\tilde{\mathbf{z}}_k^v$, estimating the probability that this segment has any correspondence in the other view. Following SuperGlue~\cite{sarlin2020superglue}, $\mathbf{S}$ is also augmented with a learnable \emph{dustbin} row and column acting as an explicit ``no-match'' slot. At inference, segment $i$ in $A$ is matched to $j^* = \arg\max_j S_{ij}$ iff $\mathrm{sigmoid}(\sigma_i^A) > 0.5$ \emph{and} $S_{ij^*}$ exceeds the dustbin score; otherwise it is left unmatched.

\subsection{Backbones: MASt3R and VGGT} \label{sec:backbones}

\textbf{MASt3R}~\cite{leroy2024mast3r} processes $I_A$, $I_B$ through a ViT-Large encoder independently, then a CroCo-style decoder~\cite{weinzaepfel2023croco} alternates self- and cross-attention between the two streams, producing patch features of shape $(H/16, W/16, 768)$; a pre-trained local-feature head additionally maps them to $24$-dim per-patch descriptors. We take this head's output, frozen, as our backbone features and pool at patch resolution. This differs from SegMASt3R, which keeps only the $768$-dim patch features frozen and learns its own segment-feature MLP on top. Either way, features are inherently \emph{pair-dependent}: $\mathbf{F}_A$ cannot be computed without $I_B$.
\textbf{VGGT}~\cite{wang2025vggt} uses a DINOv2-based ViT encoder followed by a $24$-layer Aggregator that processes patch tokens from all input views in a single shared sequence (token dim $2048$, patch size $14$); cross-view interaction emerges from self-attention. Features are \emph{pair-independent} --- they do not change when other views are added to or removed from the same forward pass --- which is what makes per-image precomputation and multi-view extensions feasible. This contrast (pairwise cross-attention vs.\ joint multi-view self-attention) dictates which heads are architecturally compatible with which backbone.

\subsection{Three matching heads} \label{sec:heads}

\paragraph{SegMASt3R+LGv2 (Fig.~\ref{fig:architecture}a).}
A two-layer MLP lifts the frozen $24$-dim MASt3R descriptors into the shared $128$-dim space, $\mathbf{z}_k = \mathrm{Linear}_{64 \to 128} \circ \mathrm{LN} \circ \mathrm{GELU} \circ \mathrm{Linear}_{24 \to 64}(\mathbf{d}_k)$. Three transformer blocks then refine the two descriptor sets, each block applying weight-shared self-attention on each view, cross-attention in both directions (with separate LayerNorms on queries and keys/values), and a position-wise FFN with expansion factor $4$. A two-layer classifier $\mathrm{Linear}(128 \to 128) \to \mathrm{ReLU} \to \mathrm{Linear}(128 \to 1)$ predicts the per-segment matchability logits. The full head adds ${\sim}800$K trainable parameters on top of MASt3R's ${\sim}1$B frozen weights. As noted above, LGv2 and Sinkhorn differ in both feature tap and matching head, so their comparison is not head-only.

\paragraph{SegVGGT-DPT (Fig.~\ref{fig:architecture}b).}
A minimal VGGT-based reference, \emph{SegVGGT (single-layer)}, pools features from the last Aggregator layer and feeds them through a small projector to $128$ dim plus the same attention stack as LGv2. Pair-independent Aggregator features make this variant precomputable per-image. Single-layer pooling, however, is spatially coarse (patch size $14$ gives a $24 \times 36$ map for $336 \times 512$ input); the deeper layers of a ViT backbone tend to be semantically rich but spatially coarse, while shallower layers retain finer spatial structure --- a well-known layer-wise trade-off~\cite{ranftl2021dpt}. We adopt the DPT~\cite{ranftl2021dpt} fusion strategy to recover boundary precision: features from Aggregator layers $\{5, 11, 17, 23\}$ are reshaped into spatial grids, projected to $256$ channels by $1{\times}1$ convolutions, and fused from deepest to shallowest through Residual Convolutional Units (pipeline notation in Supplementary Section~B). The fused map is mask-pooled (Eq.~\ref{eq:pool}) and projected to $128$ dim; the remaining attention stack and matchability head are identical to LGv2. Since fusion happens \emph{before} pooling, per-image precomputation is no longer possible. Total trainable parameters: ${\sim}10.7$M.

\paragraph{SegVGGT-DPT Joint (Fig.~\ref{fig:architecture}c).}
The pairwise heads above treat each pair in isolation; when viewpoint change is large, the two images may share very few segments, and intermediate views often bridge the gap. The Joint variant exploits this by sharing attention across $N$ views at once. Backbone, DPT fusion, pooling, and projection are applied independently to each view; the resulting per-view descriptor sets $\mathbf{z}^{(v)}$ are then concatenated into a single sequence $\mathbf{x} = [\mathbf{z}^{(1)} + \mathbf{e}_1;\; \ldots;\; \mathbf{z}^{(N)} + \mathbf{e}_N]$ with learnable per-view position embeddings $\mathbf{e}_v$ added once before the first layer. Three joint attention layers refine the sequence (self-attention + FFN, no cross-attention --- a shared sequence already lets any segment attend to any other across views); the same DoubleSoftmax scorer is then applied to any requested pair. We train with $N = 4$ and evaluate at $N \in \{2, 4, 6, 8\}$ without retraining; because joint attention treats views symmetrically and the position embeddings are sparse (only indices $0$--$3$ are exercised during training), the model generalises across $N$ without collapse.

\paragraph{Training and hyperparameters.}
We set $D = 128$, $M_{\max} = 100$ segments per image, $\tau \equiv 1$ for the MASt3R head, and a learned $\tau \leq 1$ for the VGGT heads. All models are trained on ScanNet++~\cite{yeshwanth2023scannetpp} with ground-truth correspondences from projected $3$D instance annotations ($G_{ij} = \mathbb{1}[\mathrm{id}(m_i^A) = \mathrm{id}(m_j^B)]$, where $\mathrm{id}(m)$ denotes the ground-truth instance ID of mask $m$ and $\mathbb{1}[\cdot]$ equals one when its condition is true and zero otherwise). The loss is a normalised NLL over ground-truth pairs $\mathcal{P}$, $\mathcal{L}_{\text{match}} = -\frac{1}{|\mathcal{P}|} \sum_{(i,j) \in \mathcal{P}} S_{ij}$, plus binary cross-entropy matchability with weight $\lambda = 0.3$. For the Joint model the loss is normalised per-tuple. AdamW~\cite{loshchilov2019decoupled}, learning rate $10^{-4}$, weight decay $10^{-4}$, $500$ warmup steps, cosine schedule, bf16, $2\times$ NVIDIA RTX 5090 with DDP. Supplementary Section~C collects the documented implementation settings.

%% file: sections/experiments.tex
\section{Experiments} \label{sec:experiments}

We evaluate all three matchers under a single stratified protocol, compare direct segment matching with sparse and dense aggregation in closed-loop HM3D navigation (Section~\ref{sec:downstream}), and analyse a multi-view sweep over $N$ (Section~\ref{sec:nsweep}). The full temperature-clamp ablation is in Supplementary Section~A.

\subsection{Setup} \label{sec:setup}

\paragraph{Benchmarks.}
All matchers are trained only on ScanNet++ (Section~\ref{sec:method}) and evaluated \emph{zero-shot} --- never exposed to the test domains during training --- on \textbf{Replica}~\cite{straub2019replica} (indoor) and \textbf{Virtual KITTI~2}~\cite{cabon2020vkitti2} (outdoor driving), which together probe indoor coverage, outdoor transfer, and a large appearance gap. Pairs are stratified into four angular bins by relative camera rotation ($0$--$45^\circ$, $45$--$90^\circ$, $90$--$135^\circ$, $135$--$180^\circ$); ground-truth correspondences match segments sharing an instance ID. Because both zero-shot benchmarks are rendered, Section~\ref{sec:real_images} complements them with quantitative validation on real ScanNet++ imagery and a qualitative real-robot sequence.

\paragraph{Baselines.}
We compare against five baselines: \textbf{SegMASt3R (Sinkhorn)}~\cite{sarkar2025segmastr} (our most direct prior work; frozen MASt3R patch features $\to$ learnable segment-feature MLP $\to$ Sinkhorn solver with learnable dustbin); \textbf{vizEnc+DINOv2}~\cite{oquab2023dinov2} (the original RoboHop~\cite{sarkar2024robohop} design --- here \emph{vizEnc}, short for \emph{visual encoder}, denotes a frozen appearance-only backbone whose per-mask features are pooled and matched by cosine similarity); \textbf{vizEnc+Radio} (the same recipe with frozen AM-RADIO~\cite{ranzinger2024radio} features in place of DINOv2); \textbf{LiftFeat}~\cite{liu2025liftfeat} (a recent $3$D-geometry-aware sparse local feature matcher, lifted to segment level by aggregating keypoint descriptors within each mask); and \textbf{DA3 (Sequential)}~\cite{yang2025da3} (Depth Anything 3 features pooled over masks; DA3 has no native matcher, so correspondences are obtained by greedy sequential assignment under cosine similarity until the next-best candidate's matchability falls below a threshold). We also report \emph{SegVGGT (single-layer)}, a VGGT-based head without DPT fusion, as an ablation reference: on ScanNet++ validation it reaches $0.858$ AUPRC, below SegVGGT-DPT in Table~\ref{tab:scannetpp_val}, supporting the contribution of multi-scale fusion.

\paragraph{Metrics.}
Our headline metric is \textbf{AUPRC} (area under the precision--recall curve), a threshold-free summary that rewards ranking true correspondences above false ones. We complement it with \textbf{R@1} and \textbf{R@5} (recall at $1$/$5$): the fraction of query segments whose ground-truth match is the top-ranked candidate, or within the top five. All are reported per angular bin and as a query-weighted average.

\subsection{Indoor segment matching: Replica} \label{sec:replica}

\input{tables/replica_main}

\begin{figure}[t]
  \centering
  \includegraphics[width=\linewidth]{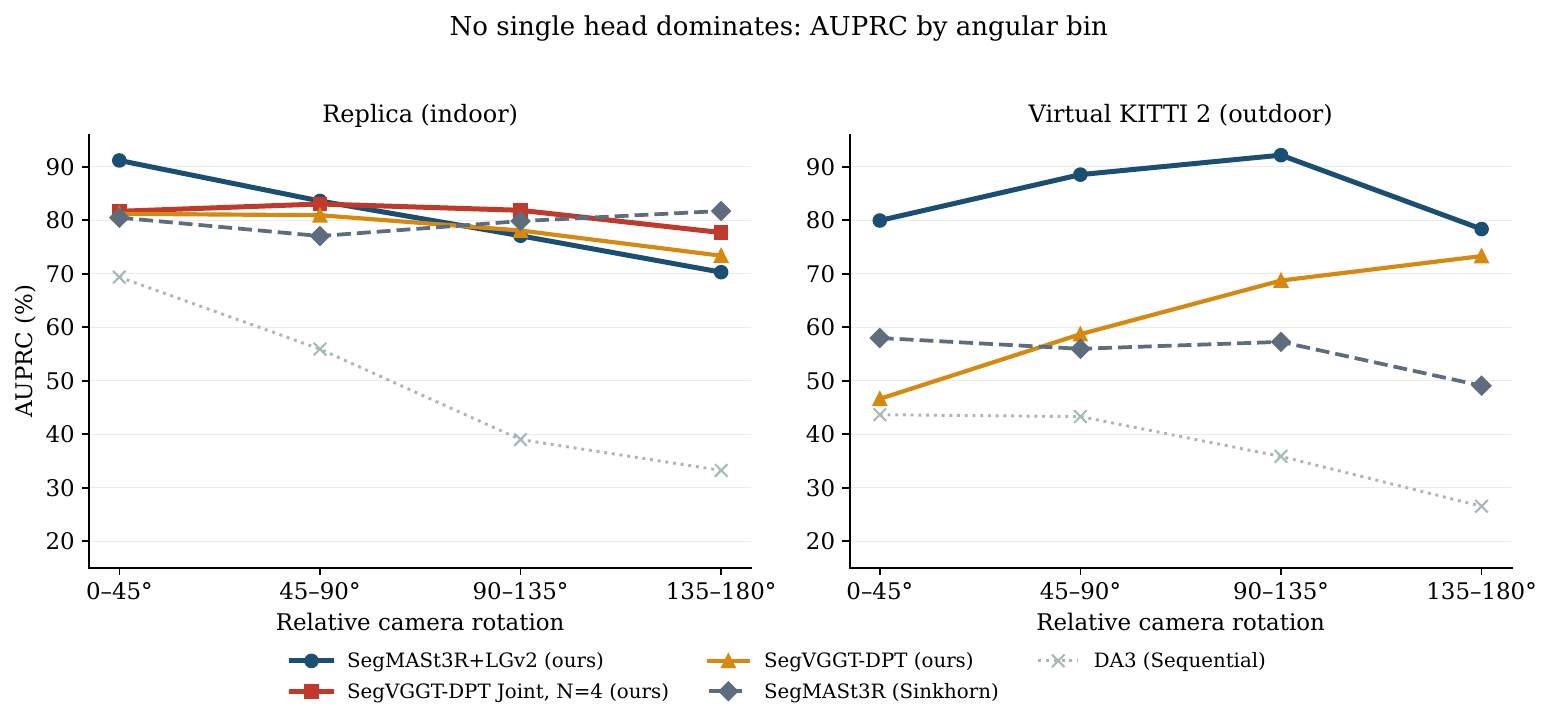}
  \caption{AUPRC per relative-rotation bin on Replica (left) and Virtual KITTI~2 (right). On Replica, LGv2 leads at narrow baselines but is overtaken by Joint (and even the Sinkhorn baseline) past $90^\circ$; on VKITTI2 LGv2 dominates every bin. The crossover is the central observation: the best head depends on the viewpoint regime.}
  \label{fig:regime}
\end{figure}

Table~\ref{tab:replica_main} reports Replica. SegMASt3R+LGv2 attains $84.48\%$ overall AUPRC, ${+}4.85$ over the Sinkhorn baseline, concentrated at narrow baselines (${+}10.7$ at $0$--$45^\circ$); but at $135$--$180^\circ$ it drops $11.4$ points \emph{below} the baseline. This reversal is informative: the learned head sharpens the regime well-represented in training yet extrapolates worse than the parameter-free matcher to low-overlap, high-rotation pairs. There, SegVGGT-DPT Joint ($N{=}4$) takes over, leading at $90$--$135^\circ$ and $135$--$180^\circ$ and beating the pairwise DPT variant by $4.3$ AUPRC at the widest bin --- isolating the multi-view contribution. The result is a regime split, not a single winner (Fig.~\ref{fig:regime}): LGv2 leads overall and at narrow baselines, while Joint leads among our heads at the two widest bins. All three are at or above the Sinkhorn baseline overall; DPT's $0.07$-point margin is not evidence of a meaningful separation.

\subsection{Outdoor zero-shot: Virtual KITTI~2} \label{sec:vkitti}

\input{tables/vkitti2_main}

LGv2 reaches $82.78\%$ AUPRC overall on Virtual KITTI~2 (Table~\ref{tab:vkitti2_main}), ${+}25.87$ over the Sinkhorn baseline and ${+}30.50$ over pairwise VGGT. MASt3R's explicit cross-view geometric coupling transfers from indoor training to outdoor driving better than VGGT's joint self-attention does, and the LGv2 head amplifies this signal rather than diluting it. Both VGGT heads exhibit an inverted driving-specific angular pattern: SegVGGT-DPT rises from $46.60\%$ to $73.29\%$ from the narrowest to the widest bin, consistent with difficulty separating near-identical segments at narrow baselines; LGv2 removes this inversion. Consistent with the indoor Replica trend, the joint multi-view head helps most exactly where the baseline is widest: SegVGGT-DPT Joint ($N{=}4$) improves the pairwise VGGT variant from $73.29$ to $76.92$ at $135$--$180^\circ$, the single bin where sharing segments across views matters most, even though its overall AUPRC is essentially tied with the pairwise variant ($52.85$ vs.\ $52.28$).

\subsection{Multi-view sweep: effect of $N$} \label{sec:nsweep}

\begin{figure}[t]
  \centering
  \includegraphics[width=\linewidth]{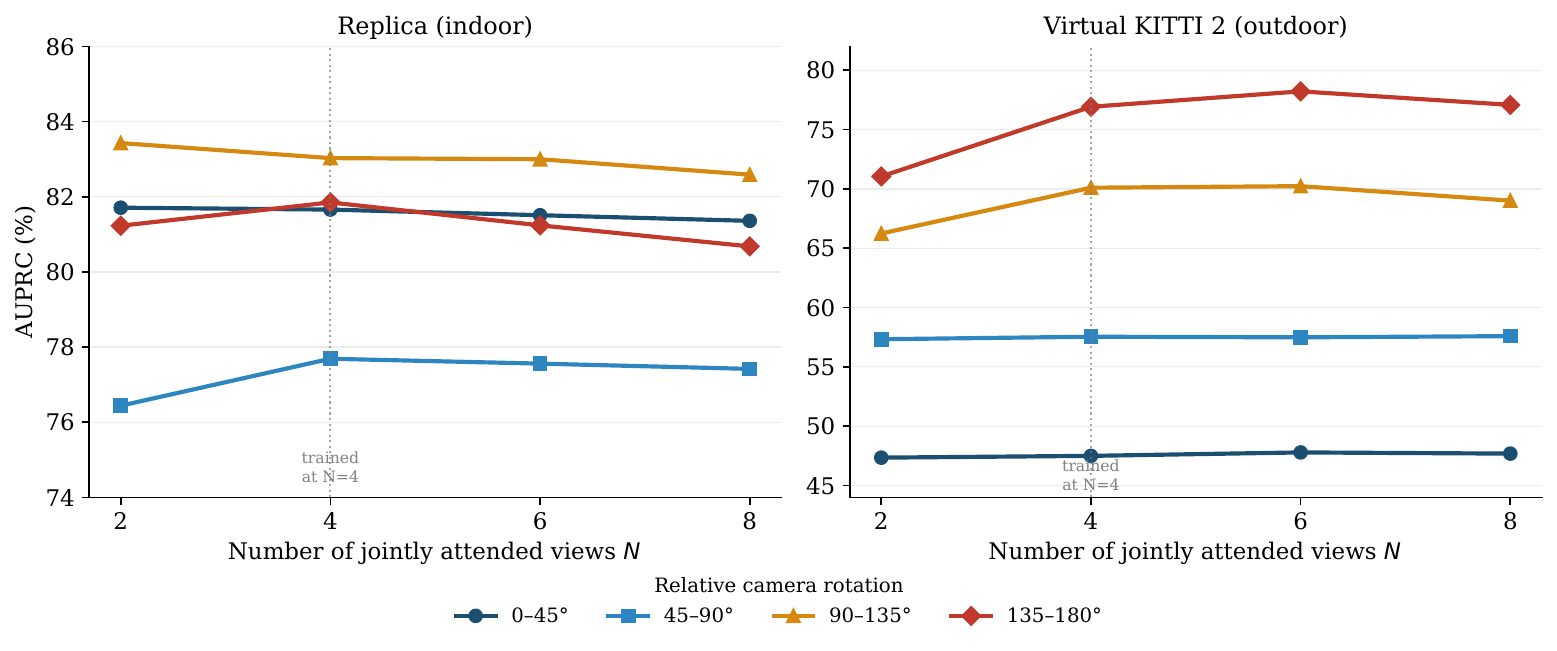}
  \caption{Effect of the number of jointly attended views $N$ on SegVGGT-DPT Joint (trained at $N{=}4$, evaluated at $N\in\{2,4,6,8\}$ without retraining). Replica (left) is flat in $N$ (dense overlap); on VKITTI2 (right) extra views help only at the widest baseline ($71.1{\to}78.2$ AUPRC at $135$--$180^\circ$) and saturate by $N{=}6$.}
  \label{fig:nsweep}
\end{figure}

Figure~\ref{fig:nsweep} plots the Joint model at $N \in \{2, 4, 6, 8\}$ per angular bin (trained with $N{=}4$, evaluated at each $N$ without retraining). On Replica $N$ has almost no effect ($81.10$--$81.57$ overall AUPRC): indoor scenes have dense per-view overlap, so extra views add little. On Virtual KITTI~2, $N$ helps specifically at wide baselines: the $135$--$180^\circ$ bin improves from $71.05$ ($N{=}2$) to $78.22$ ($N{=}6$) before saturating, while narrow bins stay flat around $47.5\%$ --- a per-pair VGGT bottleneck that adding views cannot solve. The model extrapolates to $N = 6, 8$ without collapse even though the position-embedding indices $4$--$7$ are never directly supervised, consistent with joint attention learning view-relative rather than view-absolute patterns.

\paragraph{Temperature calibration.}
Clamping the learned DoubleSoftmax temperature to $\tau \le 1$ cuts the dominant \emph{false-dustbin} errors by $36\%$ relative, re-allocating the error budget toward harder \emph{wrong-match}/\emph{false-match} cases; for this head, calibrated scoring rather than richer descriptors is the binding constraint. The full error-budget decomposition is in Supplementary Section~A.

\subsection{Evaluation on real captured imagery} \label{sec:real_images}

Replica and Virtual KITTI~2 establish zero-shot behaviour across indoor and outdoor domains, but both are rendered. We therefore add a quantitative evaluation on the ScanNet++ validation split, using real captured imagery from held-out scenes. Since all heads are trained on ScanNet++, this experiment is \emph{in-domain} rather than zero-shot; its role is to check that the gains above are not confined to rendered inputs.

\input{tables/scannetpp_val}

Table~\ref{tab:scannetpp_val} shows that all three learned heads improve AUPRC over the Sinkhorn baseline on real images. SegMASt3R+LGv2 is strongest in AUPRC ($91.0\%$) and R@5 ($98.5\%$), while the Joint head has the highest R@1 ($85.5\%$, a $0.1$-point margin over LGv2). Thus, the principal zero-shot trends do not come at the cost of in-domain real-image matching, although the small margins here should not be interpreted as evidence of real-domain generalisation.

\begin{figure}[t]
  \centering
  \includegraphics[width=0.80\linewidth]{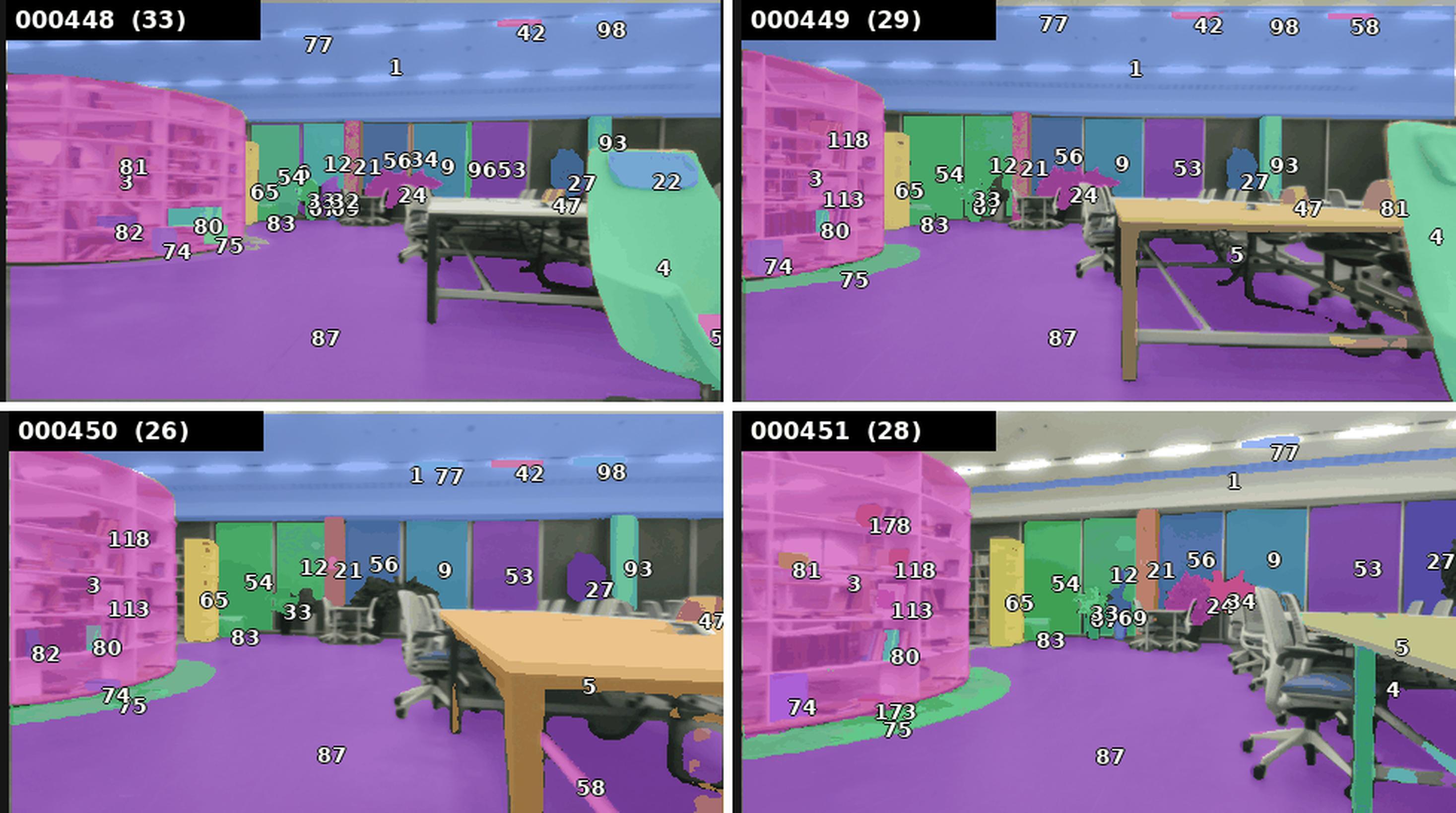}
  \caption{Joint $N{=}4$ on real robot capture without retraining: views $448$--$451$ matched in one joint pass. Colour encodes identity; headers give frame index and segment count. Only tracks spanning at least three of four views are drawn.}
  \label{fig:real_capture}
\end{figure}

Figure~\ref{fig:real_capture} provides a complementary qualitative check outside the ScanNet++ capture setup. The Joint model processes four consecutive robot views in one pass and preserves shared colour identities across camera motion even as the number and visibility of segments change between frames. Only identities observed in at least three views are displayed, so the figure illustrates multi-view consistency rather than completeness. Because this sequence has no ground-truth instance identities, it is qualitative evidence and is not included in the quantitative ranking.

\subsection{Closed-loop downstream: HM3D Instance Image Navigation} \label{sec:downstream}

Each matcher replaces the RoboHop~\cite{sarkar2024robohop} online localizer for HM3D Instance Image Navigation~\cite{krantz2023iin, ramakrishnan2021hm3d}, without retraining. The agent receives a reference image of a single object instance and must reach within $1\,\mathrm{m}$ geodesic distance (Habitat-Sim~\cite{savva2019habitat}, $250$-step cap). RoboHop's map is a topological graph of SAM segments built from a teach run; at inference, FastSAM~\cite{zhao2023fastsam} segments the current observation, the online matcher compares query segments against a $\pm 8$-frame window of the map, and a Dijkstra planner returns the next goal segment. We retain the released LightGlue-built teach-run map, isolating query-time matching as in SegMASt3R~\cite{sarkar2025segmastr}. Evaluation is on the full HM3D~v$2$ \emph{val} split --- $102$ episodes across $36$ scenes. Baselines are LightGlue~\cite{lindenberger2023lightglue} (RoboHop default, SuperPoint+LightGlue aggregated to segments) and the released SegMASt3R (Sinkhorn) checkpoint~\cite{sarkar2025segmastr}. We evaluate all three of our matchers, with Joint at $N{=}2$ and $N{=}4$, an oracle GT metric localizer as an upper bound, and RoMa~\cite{edstedt2024roma} $\to$ segment vote as a dense baseline aggregating pixel correspondences inside masks. Metrics~\cite{anderson2018spl}: Success Rate (SR), Success-weighted Path Length (SPL), Soft SPL.

Comparisons are paired: every matcher sees the same episode list with identical localizer settings, and only the online matcher differs. We test success with an exact McNemar test and per-episode SPL with a paired bootstrap ($10$k resamples). For each endpoint separately, we apply Holm correction across all $\binom{8}{2}=28$ pairwise method comparisons, including the oracle.\footnote{The seed is fixed at $42$. On a separate $10$-episode subset, repeated runs varied by roughly $\pm 6$ SPL points due to non-deterministic GPU kernels; this diagnostic is not the $102$-episode significance test.}

\input{tables/hm3d_iin}

Two separations survive correction: the oracle outperforms every matcher, and RoMa voted into masks falls below every direct segment-level matcher ($p=0.001$--$0.008$). Differences in Table~\ref{tab:hm3d_iin}'s SPL point estimates span $13.72$--$17.98$ points; this range is not a bootstrap interval. The five direct segment configurations and sparse keypoint aggregation form one cluster ($p=0.442$--$0.860$, with SPL intervals covering zero), so no ranking is claimed. The separation is specific to the tested \emph{dense} pixel-vote baseline: RoMa underperforms, whereas SuperPoint and LightGlue aggregated to segments remain in the cluster. This is not evidence against sparse aggregation or dense matchers in general.

\subsection{Discussion} \label{sec:discussion}

Two factors explain the offline regime split. First, the backbone gates the head: MASt3R's pair-dependent decoder feeds the strongest pairwise head but offers no view-scalable representation, so choosing VGGT is an enabling decision --- the price of admission to the multi-view regime --- not a matching-quality one. Second, learned capacity helps where the training data is dense and hurts outside it: LGv2 gains most at narrow baselines but falls $11$ AUPRC behind the Sinkhorn baseline at Replica $135$--$180^\circ$, having internalised the moderate-viewpoint bias of ScanNet++ co-visible pairs; the un-learned baseline is thus the safe fallback exactly where learned heads extrapolate worst.

At $102$ episodes, direct segment matchers and sparse keypoint aggregation are not statistically separated, so the offline regime split does not translate into a measurable closed-loop ordering. The supported gap is to RoMa's dense pixel-vote baseline. The real-image results in Section~\ref{sec:real_images} are complementary: ScanNet++ provides in-domain quantitative evidence, whereas the robot sequence tests multi-view identity consistency only qualitatively.

\textbf{Limitations:} (i) Training uses indoor ScanNet++ only; an outdoor source may close the VKITTI2 gap. (ii) Quantitative real-image evaluation is indoor and in-domain (ScanNet++ val, Table~\ref{tab:scannetpp_val}), not zero-shot; the in-house capture (Fig.~\ref{fig:real_capture}) is qualitative, with no ground-truth identities. Real outdoor transfer remains untested. (iii) Matching quality is bounded by the supplied masks: a segmenter-sensitivity study across SAM granularity and FastSAM is left to future work. (iv) Closed-loop evaluation is confined to simulated indoor navigation.

%% file: tables/replica_main.tex
\begin{table}[!htb]
  \centering
  \caption{Segment matching on Replica (indoor, $3{,}200$ pairs, $28{,}022$ queries), stratified by relative camera rotation. AUPRC reported per angular bin; overall AUPRC, R@1 and R@5 are query-weighted across bins. Best per column in \textbf{bold}, second best \underline{underlined}.}
  \label{tab:replica_main}
  \resizebox{\textwidth}{!}{%
  \begin{tabular}{l|cccc|c|c|c}
    \toprule
    & \multicolumn{4}{c|}{AUPRC by bin} & Overall & Overall & Overall \\
    Model & $0$--$45^\circ$ & $45$--$90^\circ$ & $90$--$135^\circ$ & $135$--$180^\circ$ & AUPRC & R@1 & R@5 \\
    \midrule
    vizEnc+Radio~\cite{ranzinger2024radio}         & 41.91             & 38.99             & 41.25             & 38.36             & 40.69             & 58.20             & 79.04 \\
    LiftFeat~\cite{liu2025liftfeat}                & 47.17             & 32.13             & 25.19             & 24.92             & 35.69             & 40.21             & 67.57 \\
    vizEnc+DINOv2~\cite{oquab2023dinov2}           & 50.78             & 47.24             & 49.49             & 50.15             & 49.48             & 59.29             & 80.29 \\
    DA3 (Sequential)~\cite{yang2025da3}            & 69.36             & 55.91             & 38.99             & 33.23             & 54.29             & 73.96             & 73.96 \\
    SegMASt3R (Sinkhorn)~\cite{sarkar2025segmastr} & 80.48             & 77.04             & \underline{79.83} & \textbf{81.72}    & 79.63             & \textbf{81.45}    & \textbf{94.21} \\
    \midrule
    SegVGGT-DPT (ours)                             & 81.20             & 80.93             & 78.07             & 73.36             & 79.70             & 74.10             & 90.11 \\
    SegVGGT-DPT Joint, $N{=}4$ (ours)              & \underline{81.66} & \underline{83.03} & \textbf{81.85}    & \underline{77.69} & \underline{81.37} & 75.64             & 90.42 \\
    SegMASt3R+LGv2 (ours)                          & \textbf{91.17}    & \textbf{83.59}    & 77.09             & 70.28             & \textbf{84.48}    & \underline{77.46} & \underline{93.63} \\
    \bottomrule
  \end{tabular}%
  }
\end{table}

%% file: tables/vkitti2_main.tex
\begin{table}[!htb]
  \centering
  \caption{Segment matching on Virtual KITTI~2 (outdoor driving, $3{,}200$ pairs, $10{,}308$ queries), under the same stratified protocol as Replica. All models trained on ScanNet++ and evaluated zero-shot. Best per column in \textbf{bold}, second best \underline{underlined}.}
  \label{tab:vkitti2_main}
  \resizebox{\textwidth}{!}{%
  \begin{tabular}{l|cccc|c|c|c}
    \toprule
    & \multicolumn{4}{c|}{AUPRC by bin} & Overall & Overall & Overall \\
    Model & $0$--$45^\circ$ & $45$--$90^\circ$ & $90$--$135^\circ$ & $135$--$180^\circ$ & AUPRC & R@1 & R@5 \\
    \midrule
    vizEnc+Radio~\cite{ranzinger2024radio}         & 29.01             & 28.29             & 23.47             & 21.12             & 27.54             & 43.38             & 66.89 \\
    vizEnc+DINOv2~\cite{oquab2023dinov2}           & 31.86             & 31.05             & 25.50             & 23.40             & 30.22             & 46.68             & 70.34 \\
    LiftFeat~\cite{liu2025liftfeat}                & 35.40             & 30.00             & 27.21             & 20.88             & 32.35             & 47.21             & 68.04 \\
    DA3 (Sequential)~\cite{yang2025da3}            & 43.64             & 43.29             & 35.84             & 26.53             & 40.67             & 53.74             & 75.40 \\
    SegMASt3R (Sinkhorn)~\cite{sarkar2025segmastr} & \underline{57.98} & 55.96             & 57.26             & 49.06             & \underline{56.91} & \underline{63.67} & \underline{81.57} \\
    \midrule
    SegVGGT-DPT (ours)                             & 46.60             & \underline{58.69} & \underline{68.72} & 73.29             & 52.28             & 42.80             & 67.27 \\
    SegVGGT-DPT Joint, $N{=}4$ (ours)              & 47.50             & 57.54             & \underline{70.10} & \underline{76.92} & 52.85             & 42.93             & 66.75 \\
    SegMASt3R+LGv2 (ours)                          & \textbf{79.94}    & \textbf{88.52}    & \textbf{92.16}    & \textbf{78.34}    & \textbf{82.78}    & \textbf{73.25}    & \textbf{96.22} \\
    \bottomrule
  \end{tabular}%
  }
\end{table}

%% file: tables/scannetpp_val.tex
\begin{table}[!htb]
  \centering
  \caption{ScanNet++ validation on real captured imagery from held-out scenes. The evaluation is indoor and in-domain, not zero-shot. AUPRC, R@1, and R@5 are percentages.}
  \label{tab:scannetpp_val}
  \begin{tabular}{lrrr}
    \toprule
    Matcher & AUPRC & R@1 & R@5 \\
    \midrule
    SegMASt3R (Sinkhorn) & $88.9$ & $83.3$ & $97.1$ \\
    SegVGGT-DPT (ours) & $89.6$ & $82.2$ & $97.6$ \\
    SegVGGT-DPT Joint (ours) & $90.6$ & $85.5$ & $97.1$ \\
    SegMASt3R+LGv2 (ours) & $91.0$ & $85.4$ & $98.5$ \\
    \bottomrule
  \end{tabular}
\end{table}

%% file: tables/hm3d_iin.tex
\begin{table}[!htb]
  \centering
  \caption{Closed-loop HM3D~v$2$ \emph{val} ($102$ episodes, $36$ scenes). Matchers share the episode list, map, masks, and localizer settings; the oracle uses ground-truth metric localization. Values are percentages, grouped by method rather than performance. The five direct segment-level configurations and sparse SuperPoint+LightGlue baseline are not statistically separated after Holm correction; no ranking is claimed.}
  \label{tab:hm3d_iin}
  \begin{tabular}{lrrr}
    \toprule
    Online matcher & SR & SPL & Soft SPL \\
    \midrule
    Oracle (GT metric localizer)                  & $78.43$ & $78.33$ & $84.29$ \\
    \midrule
    SuperPoint+LightGlue $\to$ segment vote~\cite{lindenberger2023lightglue} & $44.12$ & $44.10$ & $53.16$ \\
    SegMASt3R (Sinkhorn)~\cite{sarkar2025segmastr}   & $42.16$ & $42.36$ & $52.21$ \\
    SegMASt3R+LGv2 (ours)                         & $41.18$ & $41.17$ & $51.89$ \\
    SegVGGT-DPT (pairwise, ours)                  & $42.16$ & $42.05$ & $50.17$ \\
    SegVGGT-DPT Joint ($N{=}2$, ours)              & $45.54$ & $45.43$ & $53.60$ \\
    SegVGGT-DPT Joint ($N{=}4$, ours)              & $43.14$ & $43.03$ & $53.45$ \\
    \midrule
    RoMa $\to$ segment vote~\cite{edstedt2024roma} & $27.45$ & $27.45$ & $37.40$ \\
    \bottomrule
  \end{tabular}
\end{table}

%% file: sections/conclusions.tex
\section{Conclusion} \label{sec:conclusions}

We compared three segment-processing pipelines built on frozen geometry backbones. The result is regime-dependent rather than a single ``best'' model: LGv2 leads overall and at small-to-moderate baselines, including outdoor transfer, while Joint improves over pairwise DPT at the widest indoor baselines. The Sinkhorn baseline remains competitive in the extreme-rotation tail, consistent with learned heads' sensitivity to the moderate-viewpoint training distribution. The backbone also constrains the available heads: MASt3R supports strong pairwise matching, while VGGT enables joint multi-view attention. Closed-loop evaluation supports no ordering among direct segment matchers and sparse keypoint aggregation, but separates all of them from the tested RoMa dense-vote baseline.

Future work should test outdoor training data and extreme-rotation sampling, routing between pairwise and joint heads using deployment cues, real outdoor zero-shot transfer, and sensitivity to segmenter choice and granularity.

%% file: sections/acknowledgements.tex
\section{Acknowledgements}

The authors used Anthropic's Claude to assist with code development. All content was reviewed and verified by the authors, who take full responsibility for the final work.

%% file: main.bib
@String(CVPR  = {IEEE Conf. Comput. Vis. Pattern Recog.})

@String(ICCV  = {Int. Conf. Comput. Vis.})

@String(ECCV  = {Eur. Conf. Comput. Vis.})

@String(NeurIPS = {Adv. Neural Inform. Process. Syst.})

@String(ICLR  = {Int. Conf. Learn. Represent.})

@String(CVPRW = {IEEE Conf. Comput. Vis. Pattern Recog. Worksh.})

@String(TMLR  = {Trans. Mach. Learn Res.})

@String(CVPR  = {CVPR})

@String(ICCV  = {ICCV})

@String(ECCV  = {ECCV})

@String(NeurIPS = {NeurIPS})

@String(ICLR  = {ICLR})

@String(CVPRW = {CVPRW})

@String(TMLR  = {TMLR})

@inproceedings{sarkar2025segmastr,
  author    = {Jayanti, Rohit and Agrawal, Swayam and Garg, Vansh and Tourani, Siddharth and Khan, Muhammad Haris and Garg, Sourav and Krishna, Madhava},
  title     = {{SegMASt3R}: Geometry Grounded Segment Matching},
  booktitle = NeurIPS,
  year      = {2025}
}

@inproceedings{liu2025liftfeat,
  author    = {Liu, Yepeng and Lai, Wenpeng and Zhao, Zhou and Xiong, Yuxuan and Zhu, Jinchi and Cheng, Jun and Xu, Yongchao},
  title     = {{LiftFeat}: {3D} Geometry-Aware Local Feature Matching},
  booktitle = {Proc. IEEE Int. Conf. Robotics and Automation (ICRA)},
  year      = {2025},
  eprint    = {2505.03422},
  archivePrefix = {arXiv}
}

@article{yang2025da3,
  author    = {Lin, Haotong and Chen, Sili and Liew, Jun Hao and Chen, Donny Y. and Li, Zhenyu and Shi, Guang and Feng, Jiashi and Kang, Bingyi},
  title     = {Depth Anything 3: Recovering the Visual Space from Any Views},
  journal   = {arXiv preprint arXiv:2511.10647},
  year      = {2025}
}

@inproceedings{ranzinger2024radio,
  author    = {Ranzinger, Mike and Heinrich, Greg and Kautz, Jan and Molchanov, Pavlo},
  title     = {{AM-RADIO}: Agglomerative Vision Foundation Model Reduce All Domains Into One},
  booktitle = CVPR,
  year      = {2024}
}

@inproceedings{detone2018superpoint,
  author    = {DeTone, Daniel and Malisiewicz, Tomasz and Rabinovich, Andrew},
  title     = {{SuperPoint}: Self-Supervised Interest Point Detection and Description},
  booktitle = CVPRW,
  year      = {2018}
}

@inproceedings{sarlin2020superglue,
  author    = {Sarlin, Paul-Edouard and DeTone, Daniel and Malisiewicz, Tomasz and Rabinovich, Andrew},
  title     = {{SuperGlue}: Learning Feature Matching With Graph Neural Networks},
  booktitle = CVPR,
  year      = {2020}
}

@inproceedings{lindenberger2023lightglue,
  author    = {Lindenberger, Philipp and Sarlin, Paul-Edouard and Pollefeys, Marc},
  title     = {{LightGlue}: Local Feature Matching at Light Speed},
  booktitle = ICCV,
  year      = {2023}
}

@inproceedings{sun2021loftr,
  author    = {Sun, Jiaming and Shen, Zehong and Wang, Yuang and Bao, Hujun and Zhou, Xiaowei},
  title     = {{LoFTR}: Detector-Free Local Feature Matching With Transformers},
  booktitle = CVPR,
  year      = {2021}
}

@inproceedings{edstedt2023dkm,
  author    = {Edstedt, Johan and Athanasiadis, Ioannis and Wadenb{\"a}ck, M{\aa}rten and Felsberg, Michael},
  title     = {{DKM}: Dense Kernelized Feature Matching for Geometry Estimation},
  booktitle = CVPR,
  year      = {2023}
}

@inproceedings{edstedt2024roma,
  author    = {Edstedt, Johan and Sun, Qiyu and B{\"o}kman, Georg and Wadenb{\"a}ck, M{\aa}rten and Felsberg, Michael},
  title     = {{RoMa}: Robust Dense Feature Matching},
  booktitle = CVPR,
  year      = {2024}
}

@inproceedings{xuelun2024gim,
  author    = {Shen, Xuelun and Cai, Zhipeng and Yin, Wei and M{\"u}ller, Matthias and Li, Zijun and Wang, Kaixuan and Chen, Xiaozhi and Wang, Cheng},
  title     = {{GIM}: Learning Generalizable Image Matcher from Internet Videos},
  booktitle = ICLR,
  year      = {2024}
}

@inproceedings{wang2024dust3r,
  author    = {Wang, Shuzhe and Leroy, Vincent and Cabon, Yohann and Chidlovskii, Boris and Revaud, Jerome},
  title     = {{DUSt3R}: Geometric {3D} Vision Made Easy},
  booktitle = CVPR,
  year      = {2024}
}

@inproceedings{leroy2024mast3r,
  author    = {Leroy, Vincent and Cabon, Yohann and Revaud, Jerome},
  title     = {Grounding Image Matching in {3D} with {MASt3R}},
  booktitle = ECCV,
  year      = {2024}
}

@inproceedings{wang2025vggt,
  author    = {Wang, Jianyuan and Chen, Minghao and Karaev, Nikita and Vedaldi, Andrea and Rupprecht, Christian and Novotny, David},
  title     = {{VGGT}: Visual Geometry Grounded Transformer},
  booktitle = CVPR,
  year      = {2025}
}

@inproceedings{weinzaepfel2023croco,
  author    = {Weinzaepfel, Philippe and Leroy, Vincent and Lucas, Thomas and Br{\'e}gier, Romain and Cabon, Yohann and Arora, Vaibhav and Antsfeld, Leonid and Chidlovskii, Boris and Csurka, Gabriela and Revaud, J{\'e}r{\^o}me},
  title     = {{CroCo}: Self-Supervised Pre-training for {3D} Vision Tasks by Cross-View Completion},
  booktitle = NeurIPS,
  year      = {2022}
}

@article{oquab2023dinov2,
  author    = {Oquab, Maxime and Darcet, Timoth{\'e}e and Moutakanni, Th{\'e}o and Vo, Huy V. and Szafraniec, Marc and Khalidov, Vasil and Fernandez, Pierre and Haziza, Daniel and Massa, Francisco and El-Nouby, Alaaeldin and Assran, Mahmoud and Ballas, Nicolas and Galuba, Wojciech and Howes, Russell and Huang, Po-Yao and Li, Shang-Wen and Misra, Ishan and Rabbat, Michael and Sharma, Vasu and Synnaeve, Gabriel and Xu, Hu and J{\'e}gou, Herv{\'e} and Mairal, Julien and Labatut, Patrick and Joulin, Armand and Bojanowski, Piotr},
  title     = {{DINOv2}: Learning Robust Visual Features without Supervision},
  journal   = TMLR,
  year      = {2024}
}

@inproceedings{ranftl2021dpt,
  author    = {Ranftl, Ren{\'e} and Bochkovskiy, Alexey and Koltun, Vladlen},
  title     = {Vision Transformers for Dense Prediction},
  booktitle = ICCV,
  year      = {2021}
}

@inproceedings{kirillov2023sam,
  author    = {Kirillov, Alexander and Mintun, Eric and Ravi, Nikhila and Mao, Hanzi and Rolland, Chloe and Gustafson, Laura and Xiao, Tete and Whitehead, Spencer and Berg, Alexander C. and Lo, Wan-Yen and Doll{\'a}r, Piotr and Girshick, Ross},
  title     = {Segment Anything},
  booktitle = ICCV,
  year      = {2023}
}

@article{ravi2024sam2,
  author    = {Ravi, Nikhila and Gabeur, Valentin and Hu, Yuan-Ting and Hu, Ronghang and Ryali, Chaitanya and Ma, Tengyu and Khedr, Haitham and R{\"a}dle, Roman and Rolland, Chloe and Gustafson, Laura and Mintun, Eric and Pan, Junting and Alwala, Kalyan Vasudev and Carion, Nicolas and Wu, Chao-Yuan and Girshick, Ross and Doll{\'a}r, Piotr and Feichtenhofer, Christoph},
  title     = {{SAM} 2: Segment Anything in Images and Videos},
  journal   = {arXiv preprint arXiv:2408.00714},
  year      = {2024}
}

@article{zhao2023fastsam,
  author    = {Zhao, Xu and Ding, Wenchao and An, Yongqi and Du, Yinglong and Yu, Tao and Li, Min and Tang, Ming and Wang, Jinqiao},
  title     = {Fast Segment Anything},
  journal   = {arXiv preprint arXiv:2306.12156},
  year      = {2023}
}

@inproceedings{yeshwanth2023scannetpp,
  author    = {Yeshwanth, Chandan and Liu, Yueh-Cheng and Nie{\ss}ner, Matthias and Dai, Angela},
  title     = {{ScanNet++}: A High-Fidelity Dataset of {3D} Indoor Scenes},
  booktitle = ICCV,
  year      = {2023}
}

@article{straub2019replica,
  author    = {Straub, Julian and Whelan, Thomas and Ma, Lingni and Chen, Yufan and Wijmans, Erik and Green, Simon and Engel, Jakob J. and Mur-Artal, Raul and Ren, Carl and Verma, Shobhit and Clarkson, Anton and Yan, Mingfei and Budge, Brian and Yan, Yajie and Pan, Xiaqing and Yon, June and Zou, Yuyang and Leon, Kimberly and Carter, Nigel and Briales, Jesus and Gillingham, Tyler and Mueggler, Elias and Pesqueira, Luis and Savva, Manolis and Batra, Dhruv and Strasdat, Hauke M. and De Nardi, Renzo and Goesele, Michael and Lovegrove, Steven and Newcombe, Richard},
  title     = {The {Replica} Dataset: A Digital Replica of Indoor Spaces},
  journal   = {arXiv preprint arXiv:1906.05797},
  year      = {2019}
}

@article{cabon2020vkitti2,
  author    = {Cabon, Yohann and Murray, Naila and Humenberger, Martin},
  title     = {Virtual {KITTI} 2},
  journal   = {arXiv preprint arXiv:2001.10773},
  year      = {2020}
}

@inproceedings{savva2019habitat,
  author    = {Savva, Manolis and Kadian, Abhishek and Maksymets, Oleksandr and Zhao, Yili and Wijmans, Erik and Jain, Bhavana and Straub, Julian and Liu, Jia and Koltun, Vladimir and Malik, Jitendra and Parikh, Devi and Batra, Dhruv},
  title     = {Habitat: A Platform for Embodied {AI} Research},
  booktitle = ICCV,
  year      = {2019}
}

@inproceedings{ramakrishnan2021hm3d,
  author    = {Ramakrishnan, Santhosh K. and Gokaslan, Aaron and Wijmans, Erik and Maksymets, Oleksandr and Clegg, Alexander and Turner, John and Undersander, Eric and Galuba, Wojciech and Westbury, Andrew and Chang, Angel X. and Savva, Manolis and Zhao, Yili and Batra, Dhruv},
  title     = {{Habitat-Matterport 3D} Dataset ({HM3D}): 1000 Large-scale {3D} Environments for Embodied {AI}},
  booktitle = {Proc. NeurIPS Datasets and Benchmarks Track},
  year      = {2021}
}

@inproceedings{krantz2023iin,
  author    = {Krantz, Jacob and Gervet, Theophile and Yadav, Karmesh and Wang, Austin and Paxton, Chris and Mottaghi, Roozbeh and Batra, Dhruv and Malik, Jitendra and Lee, Stefan and Chaplot, Devendra Singh},
  title     = {Navigating to Objects Specified by Images},
  booktitle = ICCV,
  year      = {2023}
}

@inproceedings{sarkar2024robohop,
  author    = {Garg, Sourav and Rana, Krishan and Hosseinzadeh, Mehdi and Mares, Lachlan and S{\"u}nderhauf, Niko and Dayoub, Feras and Reid, Ian},
  title     = {{RoboHop}: Segment-based Topological Map Representation for Open-World Visual Navigation},
  booktitle = {Proc. IEEE Int. Conf. Robotics and Automation (ICRA)},
  year      = {2024}
}

@inproceedings{podgorski2025tango,
  author    = {Podgorski, Stefan and Garg, Sourav and Hosseinzadeh, Mehdi and Mares, Lachlan and Dayoub, Feras and Reid, Ian},
  title     = {{TANGO}: Traversability-Aware Navigation with Local Metric Control for Topological Goals},
  booktitle = {Proc. IEEE Int. Conf. Robotics and Automation (ICRA)},
  pages     = {2399--2406},
  year      = {2025}
}

@article{anderson2018spl,
  author    = {Anderson, Peter and Chang, Angel and Chaplot, Devendra Singh and Dosovitskiy, Alexey and Gupta, Saurabh and Koltun, Vladlen and Kosecka, Jana and Malik, Jitendra and Mottaghi, Roozbeh and Savva, Manolis and Zamir, Amir R.},
  title     = {On Evaluation of Embodied Navigation Agents},
  journal   = {arXiv preprint arXiv:1807.06757},
  year      = {2018}
}

@inproceedings{hughes2022hydra,
  author    = {Hughes, Nathan and Chang, Yun and Carlone, Luca},
  title     = {Hydra: A Real-time Spatial Perception System for {3D} Scene Graph Construction and Optimization},
  booktitle = {Proc. Robotics: Science and Systems (RSS)},
  year      = {2022}
}

@inproceedings{li2022sgtr,
  author    = {Li, Rongjie and Zhang, Songyang and He, Xuming},
  title     = {{SGTR}: End-to-End Scene Graph Generation With Transformer},
  booktitle = CVPR,
  year      = {2022}
}

@inproceedings{loshchilov2019decoupled,
  author    = {Loshchilov, Ilya and Hutter, Frank},
  title     = {Decoupled Weight Decay Regularization},
  booktitle = ICLR,
  year      = {2019}
}

@article{sinkhorn1967concerning,
  author    = {Sinkhorn, Richard and Knopp, Paul},
  title     = {Concerning Nonnegative Matrices and Doubly Stochastic Matrices},
  journal   = {Pacific Journal of Mathematics},
  volume    = {21},
  number    = {2},
  pages     = {343--348},
  year      = {1967}
}

@inproceedings{zhang2024mesa,
  author    = {Zhang, Yesheng and Zhao, Xu},
  title     = {{MESA}: Matching Everything by Segmenting Anything},
  booktitle = CVPR,
  pages     = {20217--20226},
  year      = {2024}
}

@article{zhang2026mesa,
  author    = {Zhang, Yesheng and Shen, Shuhan and Zhao, Xu},
  title     = {{MESA}: Effective Matching Redundancy Reduction by Semantic Area Segmentation},
  journal   = {IEEE Transactions on Pattern Analysis and Machine Intelligence},
  volume    = {48},
  number    = {4},
  pages     = {4454--4472},
  year      = {2026},
  doi       = {10.1109/TPAMI.2025.3644296}
}

@article{zhang2026a2pm,
  author    = {Zhang, Yesheng and Zhao, Xu},
  title     = {Searching from Area to Point: A Semantic Guided Framework with Geometric Consistency for Accurate Feature Matching},
  journal   = {Pattern Recognition},
  volume    = {179},
  pages     = {113920},
  year      = {2026},
  doi       = {10.1016/j.patcog.2026.113920}
}

@inproceedings{li2024masa,
  author    = {Li, Siyuan and Ke, Lei and Danelljan, Martin and Piccinelli, Luigi and Segu, Mattia and Van Gool, Luc and Yu, Fisher},
  title     = {Matching Anything by Segmenting Anything},
  booktitle = CVPR,
  pages     = {18963--18973},
  year      = {2024}
}
